\documentclass[letterpaper, 10 pt, conference]{ieeeconf}  

\IEEEoverridecommandlockouts                              

\usepackage{graphics} 
\usepackage{epsfig} 
\usepackage{amsmath} 
\usepackage{amssymb}  
\usepackage{color}
\usepackage{epstopdf}
\usepackage{mathtools}
\usepackage{makecell}
\newcommand{\PreserveBackslash}[1]{\let\temp=\\#1\let\\=\temp}
\newcolumntype{C}[1]{>{\PreserveBackslash\centering}p{#1}}
\newcolumntype{R}[1]{>{\PreserveBackslash\raggedleft}p{#1}}
\newcolumntype{L}[1]{>{\PreserveBackslash\raggedright}p{#1}}

\title{\LARGE \bf
Multi-level Contextual RNNs with Attention Model for Scene Labeling
}

\author{Heng Fan$^{1}$, Xue Mei$^{2}$, Danil Prokhorov$^{2}$ and Haibin Ling$^{1}$
\thanks{$^{1}$Heng Fan and Haibin Ling are with Computer and Information Sciences Department, Temple University, Philadelphia, PA USA.
        {\tt\small \{hengfan, hbling\}@temple.edu}}%
\thanks{$^{2}$Xue Mei and Danil Prokhorov are with Toyota Research Institute, North America, Ann Arbor, Michigan, USA.
        {\tt\small \{xue.mei, danil.prokhorov\}@toyota.com}}%
}
\pdfoutput=1
\begin{document}

\maketitle
\thispagestyle{empty}
\pagestyle{empty}

\begin{abstract}

Context in image is crucial for scene labeling while existing methods only exploit local context generated from a small surrounding area of an image patch or a pixel, by contrast long-range and global contextual information is ignored. To handle this issue, we in this work propose a novel approach for scene labeling by exploring multi-level contextual recurrent neural networks (ML-CRNNs). Specifically, we encode three kinds of contextual cues, i.e., local context, global context and image topic context in structural recurrent neural networks (RNNs) to model long-range local and global dependencies in image. In this way, our method is able to `see' the image in terms of both long-range local and holistic views, and make a more reliable inference for image labeling. Besides, we integrate the proposed contextual RNNs into hierarchical convolutional neural networks (CNNs), and exploit dependence relationships in multiple levels to provide rich spatial and semantic information. Moreover, we novelly adopt an attention model to effectively merge multiple levels and show that it outperforms average- or max-pooling fusion strategies. Extensive experiments demonstrate that the proposed approach achieves new state-of-the-art results on the CamVid, SiftFlow and Stanford-background datasets.

\end{abstract}

\section{Introduction}

\begin{figure}[!fhtb]
\centering
\begin{tabular}{@{}C{2.1cm}@{}C{2.1cm}@{}C{2.1cm}@{}C{2.1cm}@{}}
\includegraphics[width=2.05cm, height=1.45cm]{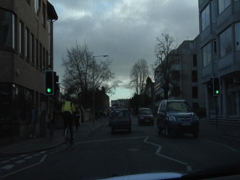} &\includegraphics[width=2.05cm, height=1.45cm]{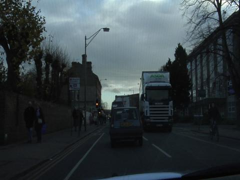} & \includegraphics[width=2.05cm, height=1.45cm]{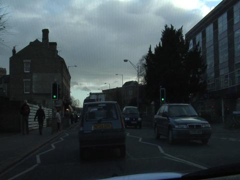} & \includegraphics[width=2.05cm, height=1.45cm]{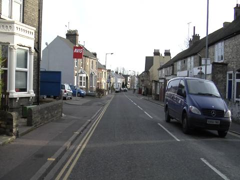}\\
\includegraphics[width=2.05cm, height=1.45cm]{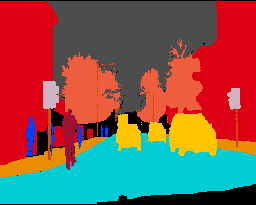} &\includegraphics[width=2.05cm, height=1.45cm]{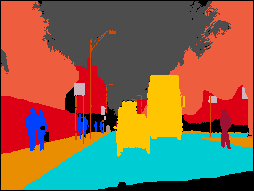} & \includegraphics[width=2.05cm, height=1.45cm]{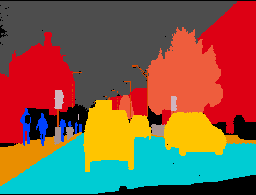} & \includegraphics[width=2.05cm, height=1.45cm]{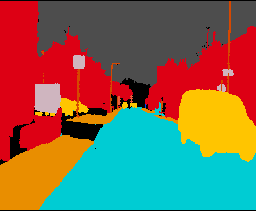}\\
\includegraphics[width=2.05cm, height=1.45cm]{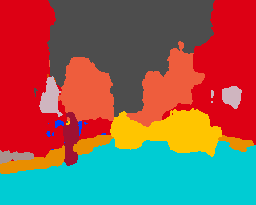} &\includegraphics[width=2.05cm, height=1.45cm]{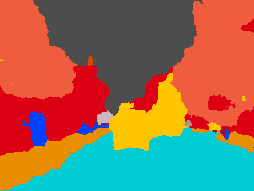} & \includegraphics[width=2.05cm, height=1.45cm]{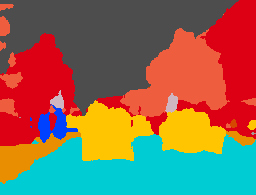} & \includegraphics[width=2.05cm, height=1.45cm]{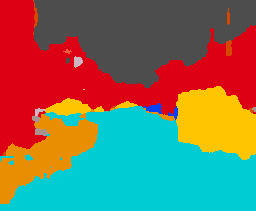}\\
\multicolumn{4}{c}{\includegraphics[width=8cm]{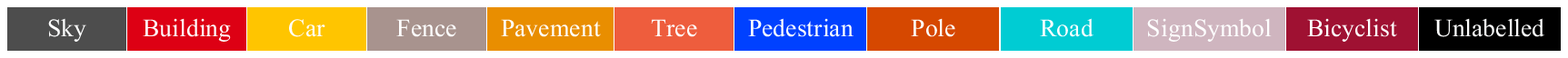}} \\
\end{tabular}
\caption{Some quantitative labeling results on CamVid. {\bf First row:} input images. {\bf Second row:} groundtruth. {\bf Third row:} our prediction labels.}\label{fig1}
\end{figure}

Scene labeling, also known as semantic segmentation, refers to assigning one of semantic classes to each pixel in an image, which plays an important role in robotics and autonomous vehicles since the robots or vehicles need to analyze and understand the environments around them. For example, the robots or vehicles must be able to discriminate building, pavement, car, pedestrian, road and so on in a traffic scene (see Figure \ref{fig1}). To address this problem, a large body of researches \cite{SRBulo,WByeon,CFarabet,CLiu,MLiang,PPinheiro,ASharma,BShuai1,BShuai,JTighe,JTighe1,JYang} have been done on scene labeling.

For image labeling, long-range context is crucial. However, existing methods mainly focus on exploiting short-range context, and thus it is prone to misclassify visually similar pixels which actually belong to different classes. For example, `sand' and 'road' pixels are hard to be distinguished with limited short-range context. However, if we consider long-range context for `sand' (i.e., `water' pixels) and `road' (i.e., `grass' pixels) pixels, their differentiations become obvious.

Recently, recurrent neural networks (RNNs) \cite{JLElman} have been successfully applied to natural language processing (NLP) \cite{AGraves2,AGraves} owing to the capability of encoding long-range contextual information among sequential data, and it only requires a limited number of network parameters. Because of these two benefits, there are some attempts to bring RNNs to the computer vision community \cite{SBell,WByeon,AGraves1,AOord,BShuai,ZZuo}. Among them, \cite{BShuai} proposes a graphical-structured RNNs to model long-range dependencies among image units.

\begin{figure*}[!htbp]
\centering
\includegraphics[width=\linewidth]{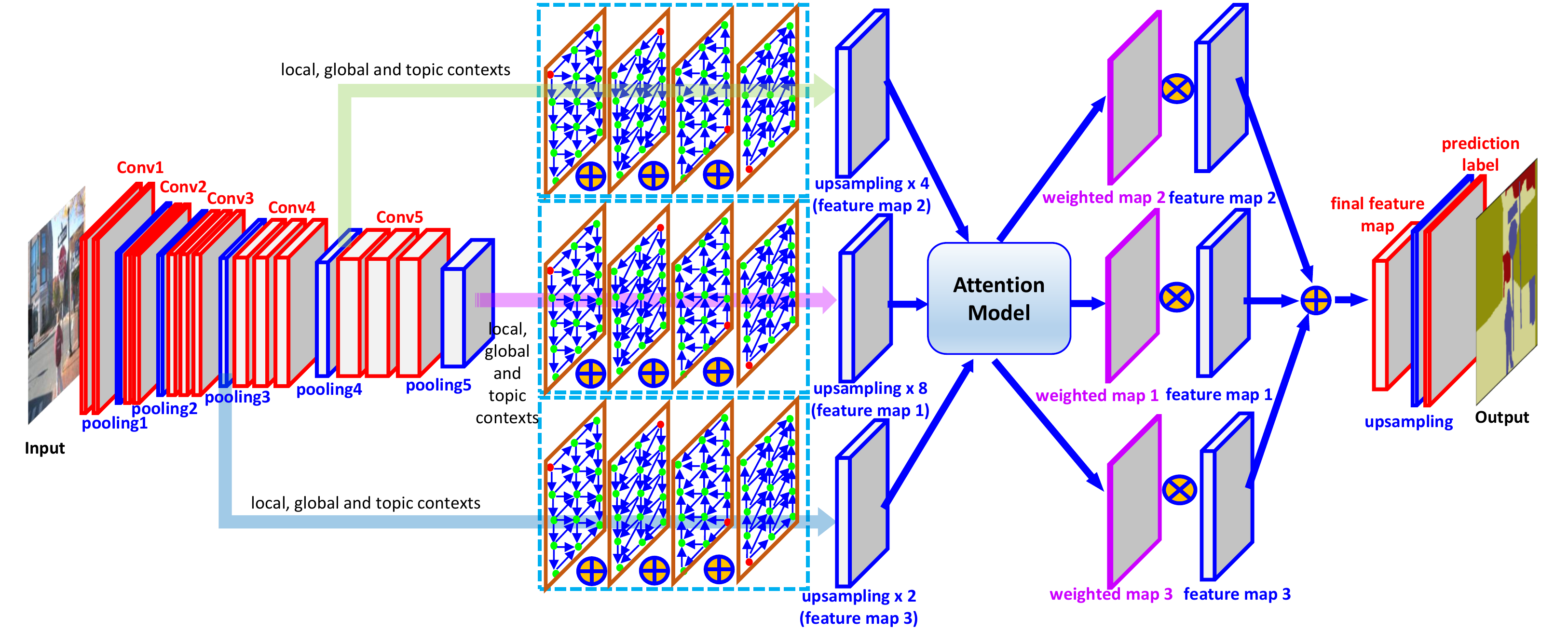}\\
\caption{Illustration of the proposed approach. We adopt CNNs to extract deep features from multiple levels, i.e., the 3$^{th}$, 4$^{th}$ and 5$^{th}$ pooling layers, which encode rich spatial and semantic information. Then multiple contextual RNNs are utilized to model dependencies in multiple levels respectively. After that, we use an attention model to effectively merge these feature maps. With the help of upsampling layers, an end-to-end network is built for image labeling. Note that the image topic features are extracted from input, and we here do not show them for conciseness.}
\end{figure*}

Inspired by this idea, we in this paper present a multi-level contextual RNNs for scene labeling. Specifically, we incorporate three kinds of contextual cues, i.e., local context, global context and image topic context in structural RNNs to model long-range local and global dependencies among image units. For local context, we consider eight neighbors for each image unit. Different from previous methods, this local context is encoded in RNNs, and the local contexts of all image units are thus connected in a structural undirected cyclic graph, which leads to the long-range local context in the whole image. However, conventional RNNs are utilized to handle sequential data and not suitable to be directly applied to structural data. We thus decompose the structural undirected cyclic graph into several directed acyclic graphs as in \cite{BShuai}. Different from \cite{BShuai}, nevertheless, we consider assigning different weights to the neighbors of each image unit because different neighbors play different roles in label inference. For example, the neighbors whose labels are the same with the image unit should play a more important role while others should be assigned with less importance. Moreover, we incorporate global and image topic contexts into RNNs which let it `see' the image in a wider view. Besides, taking the advantages of hierarchical convolutional neural networks (CNNs) into account, we integrate our contextual RNNs into CNNs, and exploit dependencies in multiple levels to provide rich spatial and semantic information. An attention model is adopted to effectively fuse these multiple levels and we show the benefits of attention model over two common fusion strategies. Integrating CNNs with RNNs, we propose an end-to-end network as shown in Figure 2. Extensive experiments on three challenging benchmarks evidence the effectiveness of our approach.

In summary, we make the following contributions:
\begin{itemize}
\item We propose the contextual RNNs which encode three kinds of contextual cues to model long-range dependencies in an image for scene labeling.
\item We exploit different dependencies in multiple levels by integrating RNNs and CNNs to provide rich spatial and semantic information for image labeling. In addition, an attention model is novelly adopted to effectively merge these multiple levels.
\item Even without any post-processing operations such as conditional random field (CRF), our method achieves new state-of-the-art results on CamVid \cite{GBrostow}, SiftFlow \cite{CLiu} and Stanford-background \cite{SGould}.
\end{itemize}

The rest of this paper is organized as follows. In Section \uppercase\expandafter{\romannumeral2}, we briefly introduce some related works of this paper. Section \uppercase\expandafter{\romannumeral3} describes the proposed approach in details. Section \uppercase\expandafter{\romannumeral4} presents experimental results, followed by conclusion in Section \uppercase\expandafter{\romannumeral5}.

\section{Related Work}

As one of the most fundamental problems in computer vision, image labeling has attracted increasing attention in recent years. Several previous non-parametric approaches try to transfer the labels of training data to the query images and perform label inference in a probabilistic graphical model (PGM). Liu {\it et al.} \cite{CLiu} propose a non-parametric image parsing method by estimating `SIFT Flow' between images, and infer the labels of pixels in a markov random field (MRF). In \cite{JTighe}, Tighe {\it et al.} introduce a superparsing method to classify superpixels by comparing $k$-nearest neighbors in a retrieval dataset, and infer their labels with MRF. Yang {\it et al.} \cite{JYang} suggest to incorporate context information to improve image retrieval and superpixel classification, and develop a four-connected pairwise MRF for semantic labeling.

The recent deep CNNs \cite{YLeCun}, which demonstrates powerfulness in extracting high-level feature representation \cite{KSimonyan}, have been successfully applied to scene labeling. In \cite{CFarabet}, Farabet {\it et al.} propose to learn hierarchical features with CNNs for scene labeling. To incorporate rich context, this method stacks surrounding contextual windows from different scales. Long {\it et al.} \cite{JLong} introduce the fully conventional networks for semantic labeling. Shuai {\it et al.} \cite{BShuai1} adopt CNNs as parametric model to learn discriminative features and integrate it with a non-parametric model to infer pixel labels. Pinheiro {\it et al.} \cite{PPinheiro} utilize CNNs in a recurrent way to model spatial dependencies in image by attaching raw input with the output of CNNs. In \cite{MLiang}, Liang {\it et al.} suggest to model the relationships among intermediate convolutional layers with RNNs for scene labeling. However, they do not consider inner structure among image units, thus the long-range dependencies in image are not captured.

\begin{figure*}[!htbp]
\centering
\includegraphics[width=\linewidth]{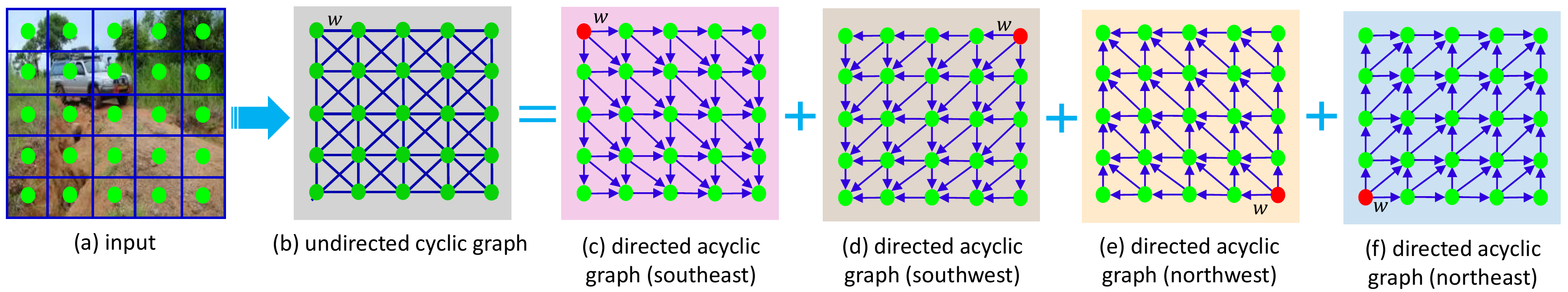}\\
\caption{Decomposition of undirected cyclic graph into four directed acyclic graphs. The green solid dots represent image units, red solid dots are start points in directed acyclic graphs, and $w$ denotes the weight. Note that in our network, the inputs for RNNs are pooling layers from CNNs, and here we just illustrate this process. Our RNNs are different from \cite{BShuai} in two aspects: (1) We assign different weights to different neighbors for each image unit, while in \cite{BShuai} the importances of different neighbors are equally weighted; (2) Our method encodes local, global and topic contexts into RNNs (see Section III-$B$), while \cite{BShuai} only considers local context.}
\end{figure*}

Recently, RNNs have drawn more and more attention in computer vision owing to the capability of capturing long-range contextual information. Oord {\it et al.} \cite{AOord} propose to model discrete probability of raw pixel values with RNNs for image completion. Graves {\it et al.} \cite{AGraves1} extend one dimensional RNNs to multi-dimensional RNNs for handwriting recognition. Based on \cite{AGraves1}, Byeon {\it et al.} \cite{WByeon} propose two-dimensional long-short term memory (LSTM) for scene parsing. This method is able to model long-range local context in image. Different from \cite{WByeon}, Zuo {\it et al.} \cite{ZZuo} propose to use hierarchical two dimensional RNNs to model spatial dependencies among image regions from multiple scales, and concatenate these dependencies for image classification. The most relevant work related to ours is \cite{BShuai}. In \cite{BShuai}, Shuai {\it et al.} use graphical RNNs to model long-range context in image and build an end-to-end trainable network by integrating with CNNs. However, our work differs from \cite{BShuai} in three aspects: (1) Considering different importances of different neighbors for each image unit, we assign different weight to each neighbor. Through this way, we propose the  weighted structural RNNs. (2) For image labeling, global and topic contexts also play  crucial roles in distinguishing pixels. We propose the contextual RNNs by incorporating local, global and topic contexts into structural RNNs. Our contextual RNNs are able to capture both long-range local and global dependencies among image units and thus `see' the entire image in a wider view. (3) To exploit rich spatial and semantic information, we integrate the contextual RNNs with CNNs and exploit various dependencies in multiple levels. An attention model is adopted to merge these multiple levels.

\section{The Proposed Approach}

In this section, we will introduce the proposed method in details. Section \uppercase\expandafter{\romannumeral3}-$A$ briefly describes the basic RNNs. Section \uppercase\expandafter{\romannumeral3}-$B$ elaborates our contextual RNNs by incorporating three contextual cues. Section \uppercase\expandafter{\romannumeral3}-$C$ illustrates the construction of multi-level contextual RNNs (ML-CRNNs) with attention model.

\subsection{Basic Recurrent Neural Networks (RNNs)}
RNNs \cite{JLElman} are developed for modeling dependencies in time sequential data. Specifically, the hidden layer $h^{(s)}$ in RNNs at time step $s$ is represented by a non-linear function over current input $x^{(s)}$ and hidden layer at previous time step $h^{(s-1)}$. The output layer $y^{(s)}$ is connected to hidden layer $h^{(s)}$.

Given an input sequence $\{x^{(s)}\}_{s=1,2,\cdots,S}$, the hidden and output layers at each time step $s$ are computed with
\begin{equation}
\begin{cases}
h^{(s)} = \phi (Ux^{(s)}+Wh^{(s-1)}+b_{h}) \\
y^{(s)} = \sigma (Vh^{(s)}+b_{y})
\end{cases}
\end{equation}
where $U$, $W$ and $V$ denote shared weight matrices between input and hidden layer, previous hidden layer and current hidden layer, and hidden layer and output layer respectively. $b_{h}$ and $b_{y}$ are two bias terms, and $\phi(\cdot)$ and $\sigma(\cdot)$ are non-linear activation functions. Since the inputs are progressively stored in hidden layers, RNNs thus can keep `memory' of the whole sequence and model long-range contextual dependencies among the sequence.

\subsection{Contextual Recurrent Neural Networks (CRNNs)}

Our CRNNs encode three contextual cues which are local, global and topic contexts. This section will introduce the incorporation of these contexts, and forward and backward operations of CRNNs.

\subsubsection{{\textbf {Local context}}}

One of our goals is to model long-range local context in image. For an image, the interactions among image units can be represented as an undirected cyclic graph (see Figure 3(b)). Due to the loopy structure of undirected cyclic graph, however, the aforementioned basic RNNs cannot be directly applied to images. To address this issue, we approximate the topology of undirected cyclic graph by the combination of several directed acyclic graphs as in \cite{BShuai}, and use variant RNNs to model long-range local context in these directed acyclic graphs as shown in Figure 3. For each directed acyclic graph, the main difference is the position of start point.

Assume that the directed acyclic graph is represented with $\mathcal{G}=\{ {\mathcal{V}, \mathcal{E}} \}$, where $\mathcal{V}=\{ v_{i} \}_{i=1,2,\cdots,N}$ denotes vertex set and $\mathcal{E}=\{ e_{ij} \}$ is the edge set in which $e_{ij}$ represents directed edge from $v_{i}$ to $v_{j}$. The structure of RNNs follows the same topology as $\mathcal{G}$. A forward propagation sequence can be seen as traversing $\mathcal{G}$ from the start point, and each vertex relies on its all predecessors. For vertex $v_{i}$, therefore, the hidden layer $h^{(v_{i})}$ is expressed as a non-linear function over current input $x^{(v_{i})}$ at $v_{i}$ and summation of hidden layers of all its predecessors. Specifically, the hidden layer $h^{(v_{i})}$ and output layer $y^{(v_{i})}$ at each $v_{i}$ are computed with
\begin{equation}
\begin{cases}
h^{(v_{i})} = \phi (Ux^{(v_{i})}+W\sum\limits_{\mathclap{{v_{j}\in{\mathcal{P}_{\mathcal{G}}(v_{i})}}}}{h^{(v_{j})}}+b_{h}) \\
y^{(v_{i})} = \sigma (Vh^{(v_{i})}+b_{y})
\end{cases}
\end{equation}
where $\mathcal{P}_{\mathcal{G}}(v_{i})$ denotes the predecessor set of $v_{i}$ in
$\mathcal{G}$. In \cite{BShuai}, the recurrent weight matrix $W$ is shared across all the predecessors of $v_{i}$. For $v_{i}$, nevertheless, different predecessors should be assigned with different weights. For example, the predecessors whose labels are the same with $v_{i}$ may be more important in inferring the label of $v_{i}$ while others play less important roles. Thus we revise Eq (2) as follows
\begin{equation}
\begin{cases}
h^{(v_{i})} = \phi (Ux^{(v_{i})}+\sum\limits_{\mathclap{{v_{j}\in{\mathcal{P}_{\mathcal{G}}(v_{i})}}}}{W^{(v_{j})}h^{(v_{j})}}+b_{h}) \\
y^{(v_{i})} = \sigma (Vh^{(v_{i})}+b_{y})
\end{cases}
\end{equation}
where $W^{(v_{j})}$ denotes the recurrent weight matrix of predecessor $v_{j}$. Through Eq (3), the RNNs are able to model long-range local context in whole image.

\subsubsection{{\textbf {Global context}}}

To further improve the ability of RNNs for pixel classification, we also consider global context in the RNNs. For the input (i.e., pooling layer in CNNs), we first partition it into 3 $\times$ 3 blocks. Then max-pooling is performed on each block. Such  partition and max-pooling result in nine feature vectors, which are concatenated as a global feature for the input. Figure 4 illustrates the extraction of global feature.

\begin{figure}[!htbp]
\centering
\includegraphics[width=8cm]{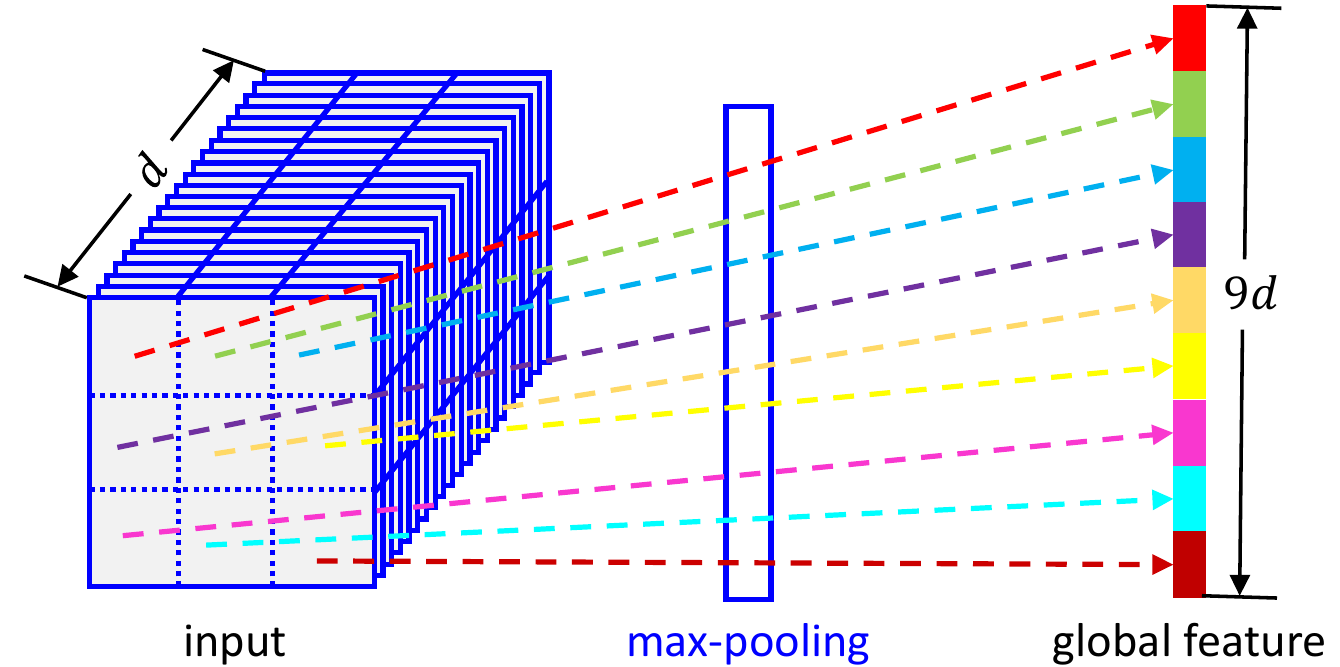}\\
\caption{Illustration of extracting global feature. The $d$ denotes the channel of input. The global feature extracted is able to capture global contextual information}
\end{figure}

Let $g=[g_{1},g_{2},\cdots,g_{9}]^{T}$ denote the global feature, where $g_{i}$ represents feature obtained by max-pooling over block $i$. To incorporate global contextual information into RNNs, we revise Eq (3) as the following
\begin{equation}
\begin{cases}
h^{(v_{i})} = \phi (Ux^{(v_{i})}+\sum\limits_{\mathclap{v_{j}\in{\mathcal{P}_{\mathcal{G}}(v_{i})}}}{W^{(v_{j})}h^{(v_{j})}}+Gg+b_{h}) \\
y^{(v_{i})} = \sigma (Vh^{(v_{i})}+b_{y})
\end{cases}
\end{equation}
where $G$ is the recurrent weight matrix for global feature $g$. Through Eq (4), the RNNs can capture both long-range local and global contextual information in the image.

\subsubsection{{\textbf {Image topic context}}}

Long-range local and global contexts can help to distinguish visually similar pixels. However, for some situations, it is still hard to classify pixels only with these two kinds of contexts. To further improve the ability of RNNs to distinguish pixels, we propose incorporating topic context information into RNNs. For instance, the `sand' pixels in Figure 5(a) and `road' pixels in Figure 5(e) are difficult to distinguish because their local and global contexts are too similar. However, if the RNNs `know' their topic features as shown in Figure (b) and (f), it will be easier to discriminate these pixels.

\begin{figure}[!htbp]
\centering
\includegraphics[width=8cm]{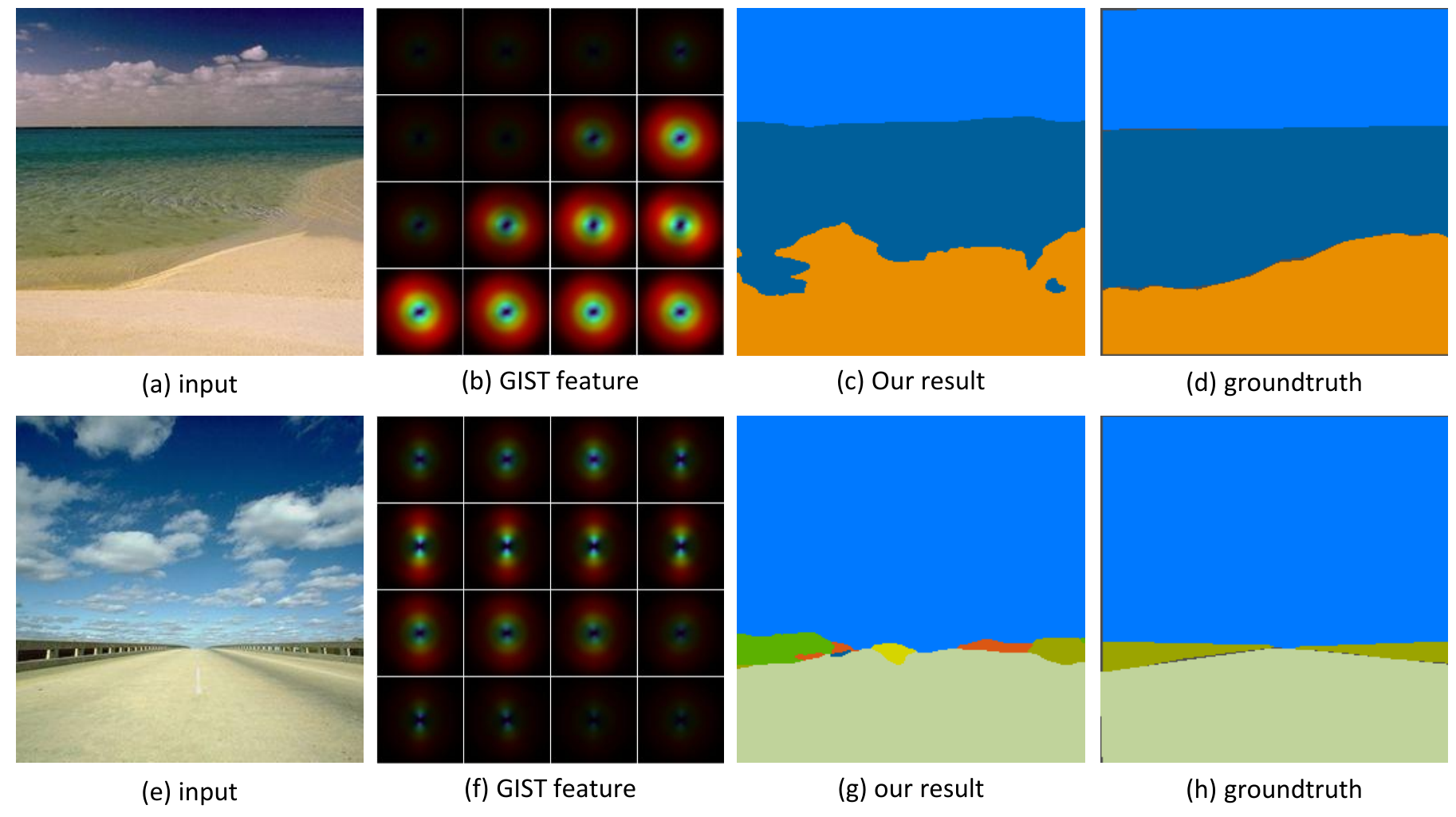}\\
\caption{Visualization of GIST feature. Images (a) and (e) are inputs, (b) and (f) are their topic features. With these topic contexts, our RNNs are able to distinguish similar pixels. Images (c) and (g) are our results, and (d) and (h) are the groundtruth.}
\end{figure}

In this paper, we adopt GIST feature \cite{AOliva} as our image topic feature. GIST feature represents the holistic abstraction of an image, and has been applied to recognition \cite{LCao1}, classification \cite{GWang}, and image retrieval \cite{SZhang}. For our network, the GIST feature is extracted from raw input image, denoted as $t$. To encode topic context, we revise Eq (4) as follows
\begin{equation}
\begin{cases}
h^{(v_{i})} = \phi (Ux^{(v_{i})}+\sum\limits_{\mathclap{v_{j}\in{\mathcal{P}_{\mathcal{G}}(v_{i})}}}{W^{(v_{j})}h^{(v_{j})}}+Gg+Tt+b_{h}) \\
y^{(v_{i})} = \sigma (Vh^{(v_{i})}+b_{y})
\end{cases}
\end{equation}
where $t$ is topic feature extracted from raw input and $T$ denotes its recurrent weight matrix. Note that topic context is different from global context. The global context encoded in global feature is still pixel-level while topic context encoded in GIST feature is image-level.

By incorporating local, global and topic contexts, our CRNNs are able to model the dependencies among image units in a wider view and thus better classify pixels.

\subsubsection{\textbf {Forward and backward operations of CRNNs}}

The CRNNs are trained via forward pass and backward propagation.
With Eq (5), we can straightforward compute the forward operation of our CRNNs.

For backward propagation, we need to calculate derivatives at each vertex in the CRNNs. For each vertex in the directed acyclic graph, it is processed in the reverse order of forward propagation sequence. In details, to compute the derivatives at vertex $v_{i}$, we need to look at the forward passes of all its successors. Let $\mathcal{S_{G}}(v_{i})$ denote the direct successor set for $v_{i}$ in $\mathcal{G}$. For each $v_{k}\in {\mathcal{S_{G}}(v_{i})}$, its hidden layer is computed with
\begin{equation}
\begin{cases}
h^{(v_{k})} = \phi (Ux^{(v_{k})}+W^{(v_{i})}h^{(v_{i})}+\mathcal{M}+Gg+Tt+b_{h}) \\
y^{(v_{k})} = \sigma (Vh^{(v_{k})}+b_{y})
\end{cases}
\end{equation}
where
\begin{equation}
\mathcal{M}=\sum\limits_{\mathclap{v_{l}\in{\mathcal{P}_{\mathcal{G}}(v_{k})}-\{ {v_{i}}\}}}{W^{(v_{l})}h^{(v_{l})}} \notag
\end{equation}
Combining Eq (5) and (6), we can see that the errors back-propagated to the hidden layer at $v_{i}$ come from two sources: directed errors from $v_{i}$ (i.e., $\frac{\partial y^{(v_{i})}}{\partial h^{(v_{i})}}$) and summation over indirected errors from all its successors
$v_{k} \in \mathcal{S_{G}}(v_{i})$ (i.e., $\sum\nolimits_{v_{k}}{\frac{\partial y^{(v_{k})}}{\partial h^{(v_{i})}}}=\sum\nolimits_{v_{k}}{\frac{\partial y^{(v_{k})}}{\partial h^{(v_{k})}}}{\frac{\partial h^{(v_{k})}}{\partial h^{(v_{i})}}}$). Therefore, the derivatives at vertex $v_{i}$ can be obtained with
\begin{equation}
\begin{cases}
\mathrm{d}h^{(v_{i})}=V^{T}\sigma{'}(y^{(v_{i})})+\sum\limits_{\mathclap{v_{k}\in{\mathcal{S_{G}}(v_{i})}}}{(W^{(v_{i})})^{T}\mathrm{d}h^{(v_{k})}\circ{\phi{'}(h^{(v_{k})})}}\\
\nabla W^{(v_{i})}=\sum\limits_{\mathclap{v_{k}\in{\mathcal{S_{G}}(v_{i})}}}\mathrm{d}h^{(v_{k})}\circ{\phi{'}(h^{(v_{k})})}(h^{(v_{i})})^{T}\\
\nabla U^{(v_{i})}=\mathrm{d}h^{(v_{i})}\circ{\phi{'}(h^{(v_{i})})}(x^{(v_{i})})^{T}\\
\nabla G^{(v_{i})}=\mathrm{d}h^{(v_{i})}\circ{\phi{'}(h^{(v_{i})})}(g)^{T}\\
\nabla T^{(v_{i})}=\mathrm{d}h^{(v_{i})}\circ{\phi{'}(h^{(v_{i})})}(t)^{T}\\
\nabla b_{h}^{(v_{i})}=\mathrm{d}h^{(v_{i})}\circ{\phi{'}(h^{(v_{i})})}\\
\nabla V^{(v_{i})}=\sigma{'}(y^{(v_{i})})(h^{(v_{i})})^{T} \\
\nabla b_{y}^{(v_{i})}=\sigma{'}(y^{(v_{i})})
\end{cases}
\end{equation}
where $\circ$ represents the Hadamard product, $\sigma{'}(\cdot)=\frac{\partial L}{\partial y(\cdot)}\frac{\partial y(\cdot)}{\partial \sigma}$ is the derivative of loss function $L$ with respect to output function $\sigma$, and $\phi{'}(\cdot)=\frac{\partial h}{\partial \phi}$. We utilize the average cross entropy loss function to compute $L$. Note that the superscript $T$ denotes transposition operation.

With Eq (5) and (7), we can perform forward and backward passes on one directed acyclic graph. In this paper, we decompose the undirected cyclic graph into four directed acyclic graphs along southeast, southwest, northwest and northeast directions as in \cite{BShuai}. Figure 3 visualizes the decomposition. Let $\mathcal{G^{U}}=\{ \mathcal{G}_{1},\mathcal{G}_{2},\mathcal{G}_{3},\mathcal{G}_{4} \}$ denote the undirected cyclic graph, where $\mathcal{G}_{1},\mathcal{G}_{2},\mathcal{G}_{3},\mathcal{G}_{4}$ represent the four directed acyclic graphs respectively. For each $\mathcal{G}_{m}$ ($m=1,2,\cdots,4$), we can get the corresponding hidden layer $h_{m}$ by performing our CRNNs. The summation of all hidden layers are fed to output layer. We use Eq (8) to express this process
\begin{equation}
\begin{cases}
h_{m}^{(v_{i})} = & \phi (U_{m}x^{(v_{i})}+\sum\limits_{\mathclap{v_{j}\in{\mathcal{P}_{\mathcal{G}_{m}}(v_{i})}}}{W_{m}^{(v_{j})}{h_{m}^{(v_{j})}}}  +Gg+Tt+b_{h_{m}}) \\
y^{(v_{i})} = &\sigma (\sum\limits_{\mathclap{\mathcal{G}_{m}\in{\mathcal{G^{U}}}}}{V_{m}h_{m}^{(v_{i})}}+b_{y})
\end{cases}
\end{equation}
where $U_{m}$, $W_{m}^{(v_{j})}$, $G$, $T$, $V_{m}$, and $b_{h_{m}}$ are matrix parameters and bias term for $\mathcal{G}_{m}$, $b_{y}$ is the bias term for final output, and $\mathcal{P}_{\mathcal{G}_{m}}(v_{i})$ denotes the predecessor set of $v_{i}$ in $\mathcal{G}_{m}$. Note that the global and topic contexts are shared across the four directed acyclic graphs. With Eq (8), we can compute loss $L$ as follows
\begin{equation}
L = -\frac{1}{N}\sum\limits_{v_{i}\in{\mathcal{G^{U}}}}\sum\limits_{j=1}^{C}{\mathrm{log}(y_{j}^{(v_{i})}Y_{j}^{(v_{i})})}
\end{equation}
where $N$ denotes the number of image units, $C$ is the number of classes, $y^{(v_{i})}$ represents class likelihood vector and $Y^{(v_{i})}$ is the binary label indicator vector for image unit at $v_{i}$. The error generated at $v_{i}$ is computed with
\begin{equation}
\nabla x^{(v_{i})} = \sum\limits_{\mathclap{\mathcal{G}_{m}\in{\mathcal{G^{U}}}}}U_{m}^{T}\mathrm{d}h_{m}^{(v_{i})}\circ{\phi{'}(h_{m}^{(v_{i})})}
\end{equation}

So far, we have introduced the proposed CRNNs and its froward and backward passes. Through encoding three contextual cues, our CRNNs are able to capture discriminative contexts in an image. In addition, our CRNNs can be seamlessly inserted into other network as intermediate layer to model dependencies among image units in last layer.

\subsection{Multi-level Contextual RNNs (ML-CRNNs) with Attention Model}

We integrate our CRNNs into CNNs to model dependencies in intermediate layers. In CNNs, different layers possess various information. The high-level layers capture more semantic information, whereas low-level layers encode more spatial information. For scene labeling, both semantic and spatial information are crucial. Therefore, we use our RNNs to exploit dependencies in multiple levels and combine them to provide rich semantic and spatial information for pixel classification. To fuse these multiple levels, average-pooling \cite{DCiresan,JDai} and max-pooling \cite{GPapandreou} are two simple and common strategies. However, different levels with different scales contain various contexts. In this paper, we propose to adopt attention model \cite{DBahdanau,LCChen} to exploit the importances of different levels. In \cite{DBahdanau}, attention model is used to softly weight the importances of words in a source sentence when predicting a target word, and \cite{LCChen} adopts attention model to weight different input data. In our work, we use attention model to weight multiple levels.

Specifically, let $\{f_{i,c}^{q}\}_{q=1,2,\cdots,Q}$ denote $Q$ feature maps of $Q$ levels, where $i$ ranges over all the spatial positions and $c\in{\{1,2,\cdots,C\}}$. Note that in our work, the feature maps from pooling layers go through CRNNs and thus have the same number of classes. All the feature maps are resized to have the same resolution via upsampling operation \cite{JLong}. We denote $z_{i,c}$ to be weighted sum of feature maps at $(i,c)$ for all levels as follows
\begin{equation}
z_{i,c}=\sum\limits_{q=1}^{Q}{\omega_{i}^{q}\cdot{f_{i,c}^{q}}}
\end{equation}
where the weight $\omega_{i}^{q}$ is calculated with
\begin{equation}
\omega_{i}^{q}=\frac{\mathrm{exp}(r_{i}^{q})}{\sum\limits_{e=1}^{Q}{\mathrm{exp}(r_{i}^{e})}}
\end{equation}
where $r_{i}^{q}$ is the feature map generated by attention model at position $i$ for level $q$. The adopted attention model consists of two convolutional layers. The first layer contains 512 filters with kernel size 3 $\times$ 3 and the second layer has $Q$ filters with kernel size 1 $\times$ 1, where $Q$ denotes the number of levels. The weight $\omega_{i}^{q}$ demonstrates the importance of feature at position $i$ in level $q$. As a consequence, the attention model is able to determine how much attention to pay to for features at different positions and levels. Besides, the attention model can be jointly trained with the networks because it allows gradient of loss function to be back-propagated \cite{DBahdanau,LCChen}.

Overall, we describe our CRNNs and attention model to exploit different dependencies in multiple levels and effectively merge them. With the help of upsamling layers, we integrate the CRNNs into CNNs and build an end-to-end network for scene labeling as shown in Figure 2.

\begin{figure*}[!fhtb]
\centering
\begin{tabular}{@{}C{2.8cm}@{}C{2.8cm}@{}C{2.8cm}@{}C{2.8cm}@{}C{2.8cm}@{}C{2.8cm}@{}}
\includegraphics[width=2.75cm, height=2.04cm]{camvid/1_ori.png} &\includegraphics[width=2.75cm, height=2.04cm]{camvid/36_ori.png} & \includegraphics[width=2.75cm, height=2.04cm]{camvid/43_ori.png} &\includegraphics[width=2.75cm, height=2.04cm]{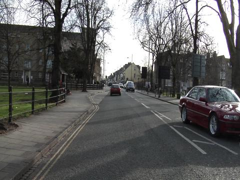} & \includegraphics[width=2.75cm, height=2.04cm]{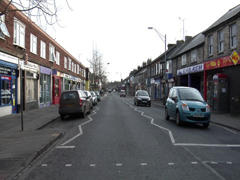} &\includegraphics[width=2.75cm, height=2.04cm]{camvid/233_ori.png} \\
\includegraphics[width=2.75cm, height=2.04cm]{camvid/1_gt.png} &\includegraphics[width=2.75cm, height=2.04cm]{camvid/36_gt.png} & \includegraphics[width=2.75cm, height=2.04cm]{camvid/43_gt.png} &\includegraphics[width=2.75cm, height=2.04cm]{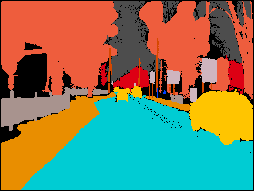} & \includegraphics[width=2.75cm, height=2.04cm]{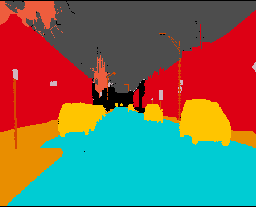} &\includegraphics[width=2.75cm, height=2.04cm]{camvid/233_gt.png} \\
\includegraphics[width=2.75cm, height=2.04cm]{camvid/1_pred.png} &\includegraphics[width=2.75cm, height=2.04cm]{camvid/36_pred.png} & \includegraphics[width=2.75cm, height=2.04cm]{camvid/43_pred.png} &\includegraphics[width=2.75cm, height=2.04cm]{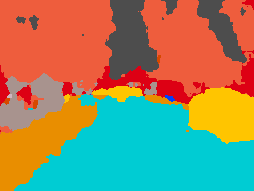} & \includegraphics[width=2.75cm, height=2.04cm]{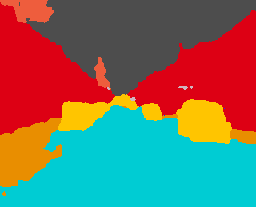} &\includegraphics[width=2.75cm, height=2.04cm]{camvid/233_pred.png} \\
\multicolumn{6}{c}{\includegraphics[width=15.6cm]{camvidlabel.eps}}\\
\end{tabular}
\caption{Quantitative labeling results on CamVid. {\bf First row:} input images. {\bf Second row:} groundtruth. {\bf Third row:} our prediction labels.}
\end{figure*}

\section{Experimental Results}

We test our approach on three benchmarks: CamVid \cite{GBrostow}, SiftFlow \cite{CLiu} and Stanford-background \cite{SGould}. Two metrics, i.e., pixel accuracy and class accuracy are adopted to evaluate the performance of our method.

\subsection{Implementation Details}

We borrow the architecture and parameters from the VGG-16 network \cite{KSimonyan} before the 5$^{th}$ pooling layer. Three independent CRNNs are utilized to model image unit dependencies in multiple levels, i.e., the 3$^{rd}$, 4$^{th}$ and 5$^{th}$ pooling layers. The dimensions of hidden layers of CRNNs are set to the same as the channels of the 3$^{rd}$, 4$^{th}$ and 5$^{th}$ pooling layers. Non-linear activation function $\phi=max(0,x)$ and $\sigma$ is {\it softmax} function. In practice, we apply $\sigma$ after final upsampling layer (see Figure 2) and use Eq (9) to compute the loss between prediction and groundtruth. The full network is trained by stochastic gradient descent (SGD) with momentum. The learning rate is initialized to be 10$^{-3}$ and decays exponentially with the rate of 0.9 after 10 epochs. The results are reported after 60 training epochs. The entire network is implemented in MATLAB using MatConvNet \cite{AVedaldi} on a single  NVIDIA GTX TITAN Z GPU with 6GB memory.

\subsection{CamVid Dataset}

CamVid is a road scene dataset which contains 701 images of day and dusk scenes \cite{GBrostow}. Each image is labelled with 11 semantic classes. We follow the usual split protocol \cite{JTighe1} (468/233) to obtain training and testing images. Table \uppercase\expandafter{\romannumeral1} demonstrates out results and comparisons with state-of-the-art methods. Figure 6 shows some qualitative labeling results of testing images in CamVid.
\renewcommand\arraystretch{1.1}
\begin{table}[!htbp]
\begin{center}
\caption{Quantitative results and comparisons on CamVid.}
\begin{tabular}{@{}C{3cm}@{}|C{2.4cm}@{}C{2.4cm}@{}}
\hline
Method  & Pixel Accuracy  & Class Accuracy \\
\hline
Bul\`o et al. \cite{SRBulo}      & 82.1\%                &  62.5\% \\
Ladick\'y et al. \cite{LLadick}  & 83.8\%                &  56.1\% \\
Shuai et al. \cite{BShuai}       & {\bf 91.6}\%          &  {\bf 78.1}\% \\
Sturgess et al. \cite{PSturgess} & 83.8\%                &  59.2\% \\
Tighe et al. \cite{JTighe}       & 78.6\%                &  43.8\% \\
Tighe et al. \cite{JTighe1}      & 83.9\%                &  62.5\% \\
Zhang et al. \cite{CZhang}       & 82.1\%                &  55.4\% \\
\hline
Our ML-CRNNs$_{avg}$        & 90.7\%                &  76.6\% \\
Our ML-CRNNs$_{max}$        & 89.9\%                &  74.9\% \\
Our ML-CRNNs$_{att}$        & {\bf 91.9}\%                &  {\bf 77.2}\% \\
\hline
\end{tabular}
\end{center}
\end{table}

From Table \uppercase\expandafter{\romannumeral1}, our method outperforms state-of-the-art approaches on pixel accuracy. However, \cite{BShuai} performs better than our method on class accuracy. We analyse two reasons accounting for this. First, \cite{BShuai} utilizes additional information of the dataset. In \cite{BShuai}, the frequency of each class is calculated. Based on the frequency, a weighting function that attends to rare class is adopted. In this way, the accuracy for non-frequent classes are phenomenally boosted. However, in real world, it is impossible to use this additional information. Second, there is only one scene involved in CamVid (i.e., the road scene). In this situation, the function of global and topic contexts is unconspicuous. Despite these two aspects, our method still obtains satisfied results owing to our multi-level CRNNs which are able to capture both semantic and spatial dependencies in image. Besides, we can also see that the adopted attention model performs better that average- or max- pooling strategies.

\subsection{SiftFlow Dataset}

The SiftFlow dataset \cite{CLiu} consists of 2688 images captured from 8 typical scenes and annotated with 33 different class labels. Following the training/testing split protocol in \cite{CLiu}, 2248 images are used for training while the rest for testing. The quantitative results and comparisons with state-of-the-art methods are listed in Table \uppercase\expandafter{\romannumeral2}. Figure 7 displays some qualitative labeling results of testing images in SiftFlow.

\begin{figure*}[!fhtb]
\centering
\begin{tabular}{@{}C{2.8cm}@{}C{2.8cm}@{}C{2.8cm}@{}C{2.8cm}@{}C{2.8cm}@{}C{2.8cm}@{}}
\includegraphics[width=2.75cm, height=2.4cm]{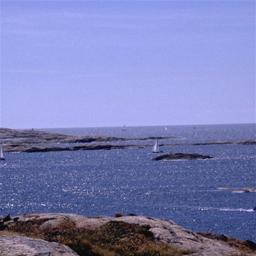} &\includegraphics[width=2.75cm, height=2.4cm]{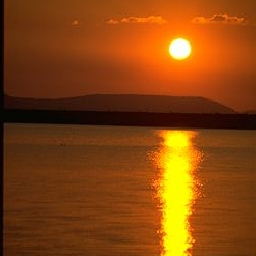} & \includegraphics[width=2.75cm, height=2.4cm]{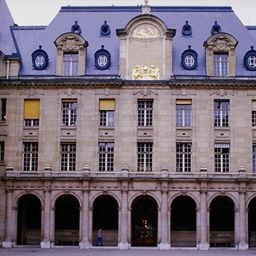} &\includegraphics[width=2.75cm, height=2.4cm]{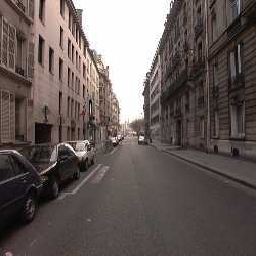} & \includegraphics[width=2.75cm, height=2.4cm]{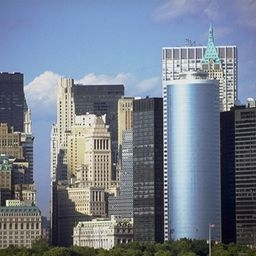} &\includegraphics[width=2.75cm, height=2.4cm]{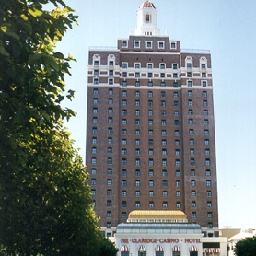} \\
\includegraphics[width=2.75cm, height=2.4cm]{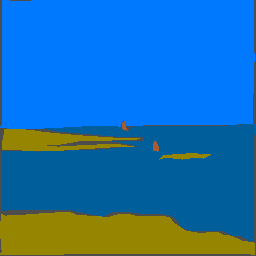} &\includegraphics[width=2.75cm, height=2.4cm]{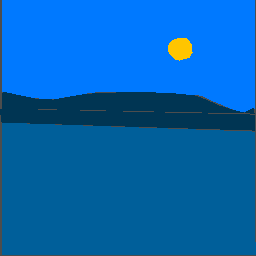} & \includegraphics[width=2.75cm, height=2.4cm]{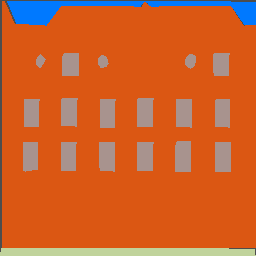} &\includegraphics[width=2.75cm, height=2.4cm]{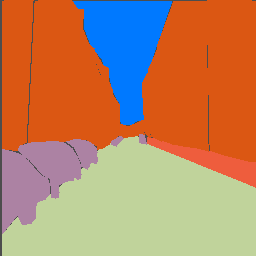} & \includegraphics[width=2.75cm, height=2.4cm]{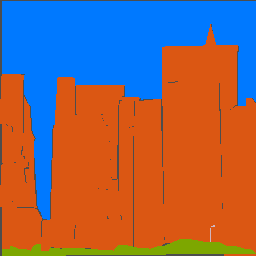} &\includegraphics[width=2.75cm, height=2.4cm]{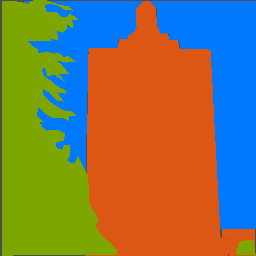} \\
\includegraphics[width=2.75cm, height=2.4cm]{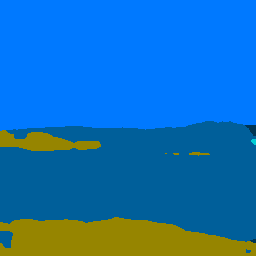} &\includegraphics[width=2.75cm, height=2.4cm]{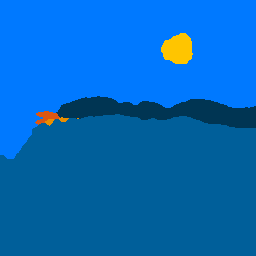} & \includegraphics[width=2.75cm, height=2.4cm]{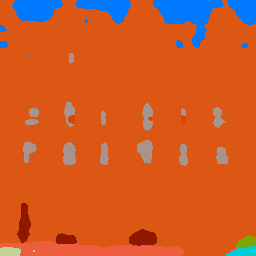} &\includegraphics[width=2.75cm, height=2.4cm]{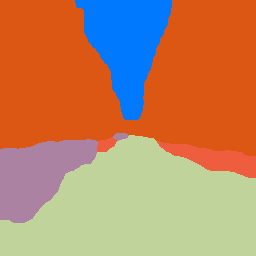} & \includegraphics[width=2.75cm, height=2.4cm]{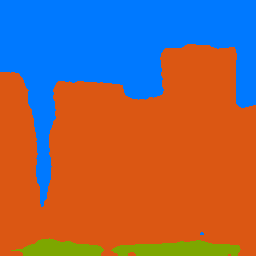} &\includegraphics[width=2.75cm, height=2.4cm]{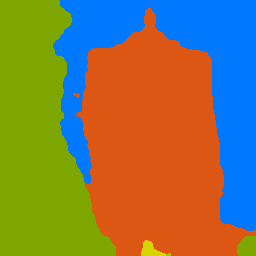} \\
\multicolumn{6}{c}{\includegraphics[width=15.6cm]{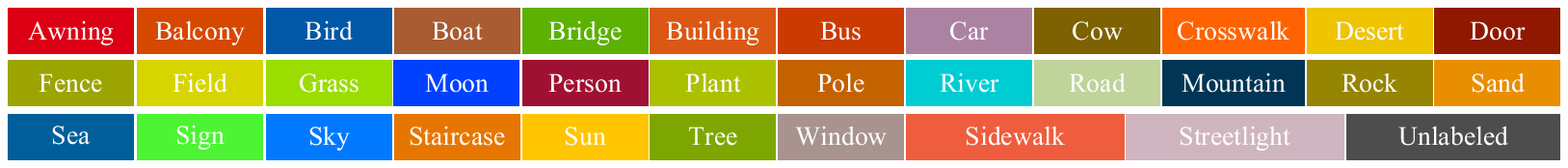}}\\
\end{tabular}
\caption{Quantitative labeling results on SiftFlow. {\bf First row:} input images. {\bf Second row:} groundtruth. {\bf Third row:} our prediction labels.}
\end{figure*}
\renewcommand\arraystretch{1.1}
\begin{table}[!htbp]
\begin{center}
\caption{Quantitative results and comparisons on SiftFlow.}
\begin{tabular}{@{}C{3cm}@{}|C{2.4cm}@{}C{2.4cm}@{}}
\hline
Method &   Pixel Accuracy  & Class Accuracy \\
\hline
Byeon et al. \cite{WByeon}        & 70.11\%                &  20.9\% \\
Farabet et al. \cite{CFarabet}    & 78.5\%                 &  29.4\%  \\
Liu et al. \cite{CLiu}            & 74.8\%                 &  -  \\
Pinheiro et al. \cite{PPinheiro}  & 77.7\%                 &  29.8\%  \\
Sharma et al. \cite{ASharma}      & 79.6\%                 &  33.6\%  \\
Shuai et al. \cite{BShuai1}       & 80.1\%                 &  39.7\%  \\
Shuai et al. \cite{BShuai}        & {\bf 85.3}\%           &  {\bf 55.7}\%  \\
Yang et al. \cite{JYang}          & 79.8\%                 &  48.7\%  \\
Liang et al. \cite{MLiang}        & 83.5\%                &  35.8\%   \\
Souly et al. \cite{NSouly}        & 80.6\%                 &  45.8\%   \\
Najafi et al. \cite{MNajafi}      & 83.1\%                 &  44.3\%   \\
\hline
Our ML-CRNNs$_{avg}$        & 85.6\%                &  55.9\% \\
Our ML-CRNNs$_{max}$        & 84.3\%                &  53.8\% \\
Our ML-CRNNs$_{att}$        & {\bf 86.9}\%                &  {\bf57.7}\% \\
\hline
\end{tabular}
\end{center}
\end{table}

From Table \uppercase\expandafter{\romannumeral2}, our proposed approach outperforms other methods on both pixel and class accuracies. Our ML-CRNNs$_{att}$ can improve the pixel accuracy from 85.3\% to 86.9\%, and the class accuracy from 55.7\% to 57.7\%. Though weighting function is adopted to improve performance in \cite{BShuai}, our method still achieves better class accuracy because our global and topic contexts are also able to help distinguish pixels which belong to rare classes.

\subsection{Stanford-background Dataset}

The Stanford-background dataset \cite{SGould} has 715 images annotated with 8 semantic classes. Following \cite{BShuai1,NSouly}, the dataset is randomly partitioned into 80\% (572 images) for training and the rest (143 images) for testing with 5-fold cross validation. As shown in Table \uppercase\expandafter{\romannumeral3}, the proposed method achieves better compared with state-of-the-art approaches. Figure 8 shows some qualitative labeling results of testing images in the Stanford-background dataset.
\renewcommand\arraystretch{1.1}
\begin{table}[!htbp]
\begin{center}
\caption{Quantitative results and comparisons on Stanford-background.}
\begin{tabular}{@{}C{3cm}@{}|C{2.4cm}@{}C{2.4cm}@{}}
\hline
Method &   Pixel Accuracy  & Class Accuracy \\
\hline
Shuai et al. \cite{BShuai1}         & 81.2\%                &   71.3\% \\
Souly et al. \cite{NSouly}          & {\bf 84.6}\%          &   {\bf 77.3}\% \\
Liang et al. \cite{MLiang}          & 83.1\%                &   74.8\% \\
Byeon et al. \cite{WByeon}          & 78.6\%               &   68.8\% \\
Pinheiro et al. \cite{PPinheiro}    &  80.2\%               &  69.9\%  \\
Farabet et al. \cite{CFarabet}      & 81.4\%                &  76.0 \%  \\
Gould et al \cite{SGould1}          & 79.3\%                &   69.4\% \\
\hline
Our ML-CRNNs$_{avg}$        & 85.7\%                &  77.1\% \\
Our ML-CRNNs$_{max}$        & 83.9\%                &  75.8\% \\
Our ML-CRNNs$_{att}$        & {\bf 87.2}\%                &  {\bf78.4}\% \\
\hline
\end{tabular}
\end{center}
\end{table}

From Table \uppercase\expandafter{\romannumeral3}, we can see the effectiveness of our ML-CRNNs$_{att}$ with attention model. Both pixel and class accuracies are significantly boosted. The pixel accuracy is improved from 84.6\% to 87.2\%, and the class accuracy is improved from 77.3\% to 78.4\%.

\begin{figure*}[!fhtb]
\centering
\begin{tabular}{@{}C{2.8cm}@{}C{2.8cm}@{}C{2.8cm}@{}C{2.8cm}@{}C{2.8cm}@{}C{2.8cm}@{}}
\includegraphics[width=2.75cm, height=2.04cm]{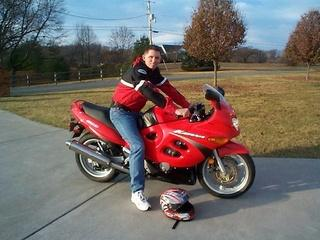} &\includegraphics[width=2.75cm, height=2.04cm]{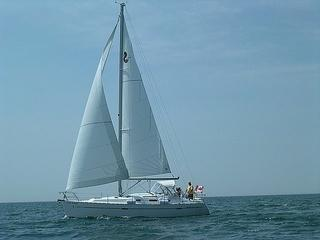} & \includegraphics[width=2.75cm, height=2.04cm]{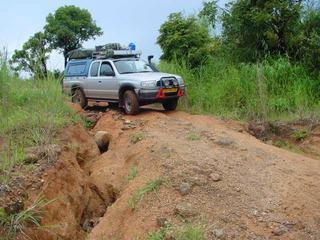} &\includegraphics[width=2.75cm, height=2.04cm]{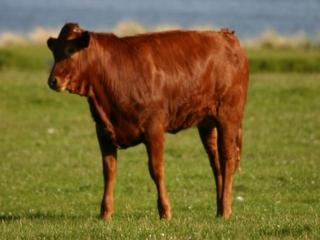} & \includegraphics[width=2.75cm, height=2.04cm]{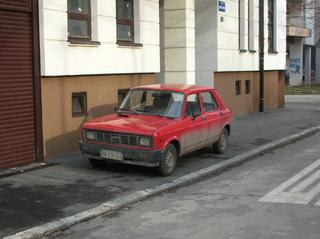} &\includegraphics[width=2.75cm, height=2.04cm]{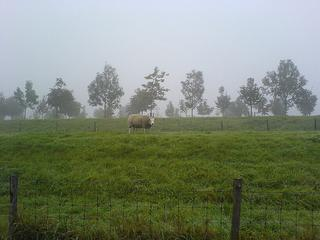} \\
\includegraphics[width=2.75cm, height=2.04cm]{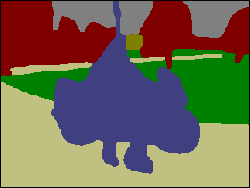} &\includegraphics[width=2.75cm, height=2.04cm]{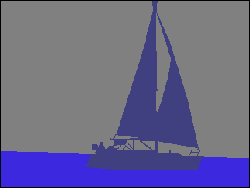} & \includegraphics[width=2.75cm, height=2.04cm]{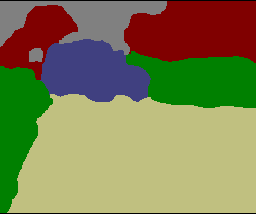} &\includegraphics[width=2.75cm, height=2.04cm]{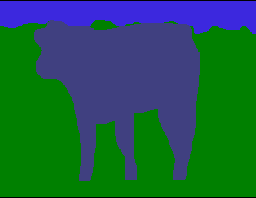} & \includegraphics[width=2.75cm, height=2.04cm]{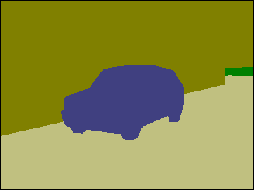} &\includegraphics[width=2.75cm, height=2.04cm]{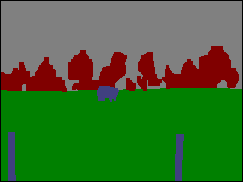} \\
\includegraphics[width=2.75cm, height=2.04cm]{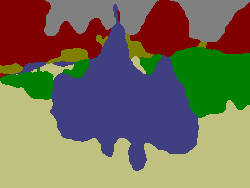} &\includegraphics[width=2.75cm, height=2.04cm]{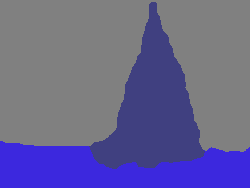} & \includegraphics[width=2.75cm, height=2.04cm]{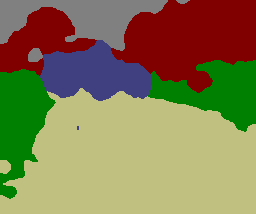} &\includegraphics[width=2.75cm, height=2.04cm]{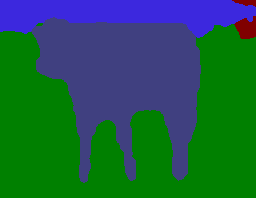} & \includegraphics[width=2.75cm, height=2.04cm]{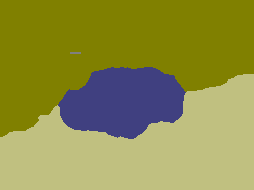} &\includegraphics[width=2.75cm, height=2.04cm]{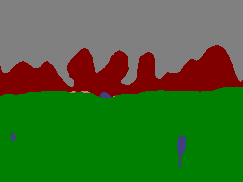} \\
\multicolumn{6}{c}{\includegraphics[width=15.6cm]{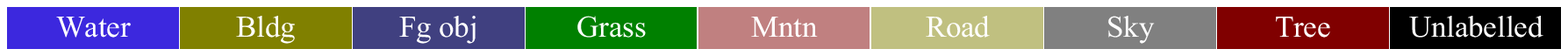}}\\
\end{tabular}
\caption{Quantitative labeling results on Stanford-background. {\bf First row:} input images. {\bf Second row:} groundtruth. {\bf Third row:} our prediction labels.}
\end{figure*}

\section{Conclusion}

In this paper, we propose the ML-CRNNs for scene labeling. We first introduce our CRNNs which are capable of capturing both long-range local, global and topic contexts in image. Moreover, to exploit different dependence relationships in multiple levels, we insert our CRNNs into CNNs to model both spatial and semantic dependencies among image units. In addition, we novelly use an attention model to learn how much attention to pay to for different levels and propose our ML-CRNNs. Extensive experiments on three challenging benchmarks demonstrate the effectiveness of our approach.



\end{document}